\documentclass[lettersize,journal]{IEEEtran}
\usepackage{graphicx}
\usepackage{multirow}
\usepackage{amsmath,amssymb,amsfonts}
\usepackage{amsthm}
\usepackage{mathrsfs}
\usepackage{xcolor}
\usepackage{textcomp}
\usepackage{manyfoot}
\usepackage{booktabs}
\usepackage{algorithm}
\usepackage{algorithmicx}
\usepackage{algpseudocode}
\usepackage{listings}

\usepackage{verbatim}
\usepackage[switch]{lineno}

\begin{document}

\title{Gradient events: improved acquisition of visual information in event cameras}
\author{Eero Lehtonen, Tuomo Komulainen, Ari Paasio, Mika Laiho\\ Kovilta, Finland
\thanks{This paper was produced by Kovilta Oy. They are in Turku, Finland.}
\thanks{Correspondence: firstname.lastname at kovilta.fi}%
}

\markboth{}%
{Lehtonen \MakeLowercase{\textit{et al.}}: Gradient events: improved acquisition of visual information in event cameras}

\maketitle

\begin{abstract}
The current event cameras are bio-inspired sensors that respond to brightness changes in the scene asynchronously and independently for every pixel, and transmit these changes as ternary event streams. Event cameras have several benefits over conventional digital cameras, such as significantly higher temporal resolution and pixel bandwidth resulting in reduced motion blur, and very high dynamic range. However, they also introduce challenges such as the difficulty of applying existing computer vision algorithms to the output event streams, and the flood of uninformative events in the presence of oscillating light sources. Here we propose a new type of event, the \emph{gradient event}, which benefits from the same properties as a conventional brightness event, but which is by design much less sensitive to oscillating light sources, and which enables considerably better grayscale frame reconstruction. We show that the gradient event -based video reconstruction outperforms existing state-of-the-art brightness event -based methods by a significant margin, when evaluated on publicly available event-to-video datasets. Our results show how gradient information can be used to significantly improve the acquisition of visual information by an event camera.
\end{abstract}


\maketitle

\begin{figure}[!b]
    \centering
    \includegraphics[width=0.7\linewidth]{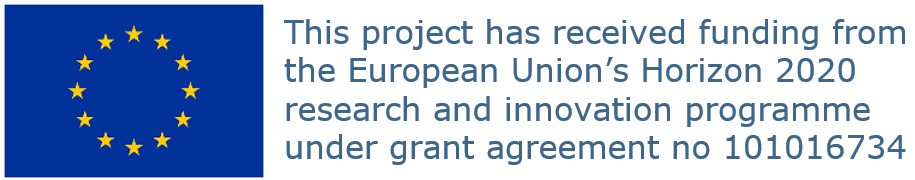}
\end{figure}

\section{Introduction}\label{sec:introduction}
The development of event cameras started with Mahowald's and Mead's~\emph{silicon retina}~\cite{mahowald-scientific-american}, which was an electrical emulation of the top three layers of the vertebrate retina, including photoreceptors, horizontal cells and bipolar cells. The silicon retina's output was an analog voltage proportional to the difference between a center pixel intensity and a weighted average of the intensities of neighboring pixels, where the pixels were arranged in a hexagonal lattice. Subsequently, all five layers of the retina were emulated by the silicon chip~\cite{boahen-silicon-retina}, which outputs spike trains that mimic (to some degree of accuracy) the outputs of ON- and OFF-center retinal ganglion cells. 

Although heavily influenced by these seminal works, in the past years, the term ``event camera'' has become practically synonymous with the representation of (logarithmic) brightness changes in pixels --- cf.~\cite{dvs128, davis, atis} for some prominent realizations --- where the change is communicated as a ternary event typically using the so-called address-event representation~\cite{event-based-vision-a-survey}. Here we note that a recent publication~\cite{tobi_center_surround} revisits the idea of a center-surround event camera, proposes a compact design for it, and describes how it avoids some of the drawbacks of conventional event cameras such as the storm of uninformative events due to flicker in many artificial lighting systems (\emph{e.g.}, sodium-vapor lamps, LEDs, and fluorescent lamps).

In this work we call the events produced by a conventional brightness-sensing event cameras~\emph{brightness events}, and will propose a novel~\emph{gradient event} that could be implemented in future generations of event cameras. To justify the used terminology, we define the concept of an event as follows:
\begin{enumerate}
    \item an event takes typically a ternary value (\emph{i.e.}, value from the set $\{-1,0,1\}$)
    \item the value $0$ means that the input signal has remained approximately constant
    \item a significant change in the input signal yields immediately an event. Here the input and output signals have some defined bandwidths, and the word immediately is interpreted with respect to these bandwidths
    \item events can be computed by independent units based on local operations
\end{enumerate}
Conventional brightness events --- computed by monitoring the changes of the (logarithmic) intensity at each pixel --- satisfy these requirements 1--4, and as we will show, so does the proposed gradient event.

A gradient camera (a conventional imager, not an event camera), which estimates local image gradients at each pixel, was originally proposed in~\cite{why-i-want-a-gradient-camera}. Several benefits of measuring gradients instead of intensities were identified, including the possibility to acquire high dynamic range (HDR) images while keeping the local analog-to-digital dynamic range small (useful when quantizing fine details), smoother quantization noise in the resulting grayscale images, and correctable saturation. Reconstruction of relatively high-quality grayscale images was possible even with two-bit quantization of gradient images. An implementation of a binary gradient camera is presented in~\cite{NVIDIA-retrieving-gray-level-from-binary-sensor}, where the use case of gesture detection was investigated. A follow-up paper~\cite{NVIDIA-reconstructing-intensity-images-from-binary-spatial-gradient-cameras} described a neural network reconstruction of grayscale images from binary gradients.

We adopt the ideas presented in~\cite{why-i-want-a-gradient-camera} to the event camera domain. In particular, we describe how ternary quantization of gradients with position-dependent thresholding and temporal difference -type encoding can be used to construct a new type of event with better acquisition of visual information as compared to the conventional brightness event. To assess the quality of the acquisition of visual information, we use a Poisson-solver based method to reconstruct grayscale images from the gradient events and compare these to brightness event -based reconstructions.

Reconstruction of grayscale frames from brightness events has mainly concentrated (\emph{e.g.}~\cite{E2VID, Firenet, HQF, SPADE-E2VID, SSL-E2VID, ET-Net, HyperE2VID}) on using various neural network architectures which take tensors of brightness events as inputs. These network architectures typically have from tens of thousands to millions of parameters, and may result in loss of temporal resolution in the reconstruction output, as the duration of the input tensors is typically significantly longer than the temporal resolution of the individual brightness events. In comparison, the gradient event reconstruction method presented in this work uses only three adjustable parameters and is based only on the latest values of gradient events thus resulting in no loss of temporal resolution. We note that the recurrent neural network -based method presented in~\cite{HyperE2VID} can achieve very high temporal resolution (in the order of to thousands of frames per second), and in future it would be interesting to quantitatively compare the high frame-rate reconstruction capabilities between~\cite{HyperE2VID} and the proposed gradient event -based method.

A majority of the results presented in this work concentrate on reconstruction quality. Although many downstream applications (\emph{e.g.}, corner detection, object tracking, optical flow, and object detection) of gradient events might not require reconstruction at all, we justify the emphasis on reconstruction as a way to estimate the amount of information retained in the seemingly very lossy ternary quantization. 

\section{Methods}
\subsection{From gradient images to ternary gradients}\label{subsec:ternary_gradients}
Let us denote grayscale images by $I(x,y) \in [0,1]_{H\times W}$, where $x = 0, \ldots, W - 1$ and $y = 0, \ldots, H - 1$. Here $W$ and $H$ are the width and the height of the image, respectively.

We define the gradient images $G_X\in [-1,1]_{H\times W}$ and $G_Y\in [-1,1]_{H\times W}$, in horizontal and in vertical directions, respectively, as
\begin{equation}\label{eq:gx}
    G_X(x, y) = I(x + 1, y) - I(x, y) \ \text{for $x = 0, \ldots, W - 2$},
\end{equation}
and
\begin{equation}\label{eq:gy}
    G_Y(x, y) = I(x, y + 1) - I(x, y) \ \text{for $y = 0, \ldots, H - 2$},
\end{equation}
where we set $G_X(W-1,y) = G_Y(x,H-1) = 0$ for $y=0,\ldots,H-1$ and $x=0,\ldots,W-1$.

\emph{Ternary gradients} are obtained by thresholding the gradient images $G_X$ and $G_Y$, where the threshold value depends on the pixel position. For this, we use a set of $n$ gradient thresholds denoted by $\{t_0, \ldots, t_{n-1}\}$, where each threshold $t_i > 0$. Now, a threshold matrix $\Theta$ of width $W$ and height $H$ is defined as
\begin{equation}
    \Theta(x,y) = \theta_{xy}, \ \text{where $\theta_{xy} \in \{t_0, \ldots, t_{n-1}\}$ for all $x$ and $y$.}
\end{equation}
In the examples and results presented in this work, we use three thresholds $\{t_0, t_1, t_2\}$ and define
\begin{equation}
    \Theta(x, y) = t_i, \ \text{where $i \equiv x + y \pmod{3}$.}
\end{equation}
For example, the upper left corner of the threshold matrix is then organized as
\begin{equation}
\Theta(x,y) = \begin{pmatrix}
    t_0 & t_1 & t_2 & t_0 & \ldots \\
    t_1 & t_2 & t_0 & t_1 & \ldots \\
    t_2 & t_0 & t_1 & t_2 & \ldots \\
    \vdots & & & & \ddots
\end{pmatrix}.
\end{equation}

The threshold matrix $\Theta$ allows us to use \emph{position-dependent} threshold values in quantizing the gradients. The ternary gradients $T_X$ and $T_Y$ are defined as
\begin{equation}
    T_X(x,y) = \begin{cases} 0, \  \text{if $|G_X(x,y)| < \Theta(x,y)$,} \\ \textrm{sgn}(G_X(x,y)), \ \text{otherwise,} \end{cases}
\end{equation}
and
\begin{equation}
    T_Y(x,y) = \begin{cases} 0, \ \text{if $|G_Y(x,y)| < \Theta(x,y)$,} \\ \textrm{sgn}(G_Y(x,y)), \ \text{otherwise,} \end{cases}
\end{equation}
where the sign function $\textrm{sgn}(\cdot)$ equals $+1$ for positive input values, $-1$ for negative input values. The corresponding \emph{quantized gradients} $\hat{G}_X$ and $\hat{G}_Y$ are defined as
\begin{equation}
    \hat{G}_X(x,y) = \begin{cases} 0, \ \text{if $|G_X(x,y)| < \Theta(x,y),$} \\ \textrm{sgn}(G_X(x,y))\cdot\Theta(x,y), \ \text{otherwise,} \end{cases}
\end{equation}
and
\begin{equation}
    \hat{G}_Y(x,y) = \begin{cases} 0, \ \text{if $|G_Y(x,y)| < \Theta(x,y)$,} \\ \textrm{sgn}(G_Y(x,y))\cdot\Theta(x,y), \ \text{otherwise.} \end{cases}
\end{equation}

As can be seen, the quantized gradients can be obtained as a element-wise product between the ternary gradients and the threshold matrix, for example $\hat{G}_X(x,y) = T_X(x,y) \circ \Theta(x,y)$, where $\circ$ denotes element-wise multiplication. This means that if the threshold matrix $\Theta(x,y)$ is known by both sender and receiver, it is enough to transmit only the ternary gradients $T(x,y)$ for the receiver to decode the values $\hat{G}(x,y)$. 

\begin{figure*}
\centering
\includegraphics[width=\linewidth]{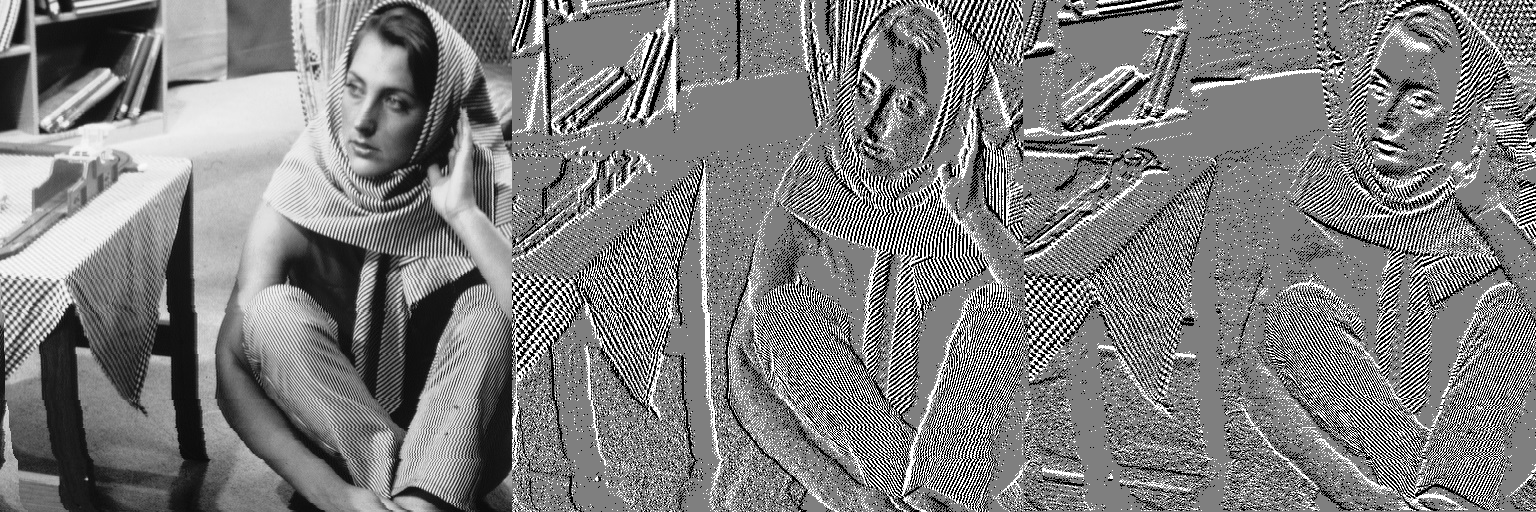}
\caption{Ternary gradients using the position dependent thresholds $\{t_0 = 4/255, \ t_1 = 8/255, \ t_2 = 16/255\}$. Left inset: original image. Middle inset: horizontal ternary gradient $T_X(x,y)$. Right inset: vertical ternary gradient $T_Y(x,y)$. In the middle and right insets, ternary gradients equal to $+1$ are represented by white pixels, and ternary gradients equal to $-1$ are represented by black pixels. Gray pixels denote positions where the ternary gradient equals zero.}\label{fig:gradient_events}
\end{figure*}

Fig.~\ref{fig:gradient_events} illustrates ternary gradients computed from an example image, with position-dependent quantization of gradients using the quantization thresholds $\{t_0 = 4/255, \ t_1 = 8/255, \ t_2 = 16/255\}$ (in this work the ground truth images are 8-bit grayscale images which are normalized to the range $[0,1]$). 

\subsection{Gradient events}
We define gradient events as ternary events which describe --- without loss of information --- how ternary gradients of an image change with time; a rule for computing the value of a gradient event from previous and current ternary gradients is presented in Table~\ref{table:gradient_event}. Computation of ternary gradients, and therefore gradient events can be performed independently and asynchronously by each pixel, provided they have access to the intensity values measured by their neighboring pixels.

Notice that this is just one possible rule for computing the gradient event, the other seven possible lossless rules can be obtained by multiplying any of the rows of the $3\times 3$ lower-right sub-matrix by $-1$. In order to satisfy the requirements 1 and 2 defined in Section~\ref{sec:introduction}, all the elements of this matrix must belong to $\{-1, 0, 1\}$, and the diagonal elements corresponding to unchanged value of the ternary gradient must equal $0$. Furthermore, in order to have lossless encoding, each row must have \emph{all} of the values $\{-1, 0, 1\}$ in some order.

\begin{table}
\caption{\label{table:gradient_event}Gradient event $E(x,y)$ as a function of previous and current ternary gradients $T_\textrm{prev}(x,y)$ and $T(x,y)$. Notice that this defines a lossless code: knowing $T_\textrm{prev}(x,y)$ and $E(x,y)$ allows to compute the value $T(x,y)$.}
\centering
\begin{tabular}{r|rrr}
\toprule
&  \multicolumn{3}{c}{$T(x,y)$} \\
$T_\textrm{prev}(x,y)$ & -1 & 0 & 1 \\
\midrule
-1 & 0 & -1 & 1 \\
0 & -1 & 0 & 1 \\
1 & -1 & 1 & 0 \\
\toprule
\end{tabular}
\end{table}

\subsection{Resolution compression}
A conventional event camera outputs data whose spatial resolution $H \times W$ corresponds to the resolution of the pixel array. In contrast to this, a gradient event camera outputs two streams of events (corresponding to the horizontal and the vertical gradients of an image), $E_X$ and $E_Y$, both of which have the same spatial resolution $H\times W$. To make the comparison against reconstruction methods using brightness events in Section~\ref{sec:results} fair, the gradient images can be resized to half of the original resolution before ternary quantization to make the overall number of values equal to $H\cdot W$: the horizontal gradient image is resized by half in the horizontal direction, and the vertical gradient image is resized by half in the vertical direction. In this work we call this operation \emph{resolution compression}, and denote the corresponding approximate gradients by $\hat{G}_X^\textrm{RC}(x,y)$ and $\hat{G}_Y^\textrm{RC}(x,y)$.

Mathematically, resolution compressed gradient images $G_X^\textrm{RC}$ and $G_Y^\textrm{RC}$ are computed from the gradient image $G_X$ and $G_Y$ as
\begin{align}\label{eq:gx_compressed}
\begin{aligned}
G_X^\textrm{RC}(x,y) &= G_X^\textrm{RC}(x+1,y) \\
&\equiv (G_X(x, y) + G_X(x+1, y)) / 2,
\end{aligned}
\end{align}
for even $x = 0, 2, \ldots, W - 2$, and 
\begin{align}\label{eq:gy_compressed}
\begin{aligned}
G_Y^\textrm{RC}(x,y) &= G_Y^\textrm{RC}(x,y+1) \\
&\equiv (G_Y(x, y) + G_Y(x, y+1)) / 2,
\end{aligned}
\end{align}
for even $y = 0, 2, \ldots, H - 2$. Here we assume that the height $H$ and the width $W$ of the image are both even numbers. In other words, horizontal gradients are averaged in the horizontal direction and vertical gradients are averaged in the vertical direction. Now, since half of the gradient values are duplicates (and need not be transmitted as gradient events), it follows that the total number of possibly unique gradient values equals $H\cdot W$. Quantization of gradients and the computation of ternary gradients can be performed as described in Section~\ref{subsec:ternary_gradients}, noting that quantization is performed only once per a pair of duplicate values defined in~\eqref{eq:gx_compressed} and~\eqref{eq:gy_compressed}.

\subsection{Reconstruction of grayscale images}\label{subsec:reconstruction}
From the ternary gradients it is possible to reconstruct a grayscale image by solving an approximation of the Poisson equation, which relates the two-dimensional discrete Laplacian to the grayscale image. The reconstruction process can be hence divided into two parts: the computation of an approximation of the discrete Laplacian, and the computation of the solution of the Poisson's equation.

\subsubsection{Computing an approximation of the discrete Laplacian}
The two-dimensional \emph{discrete Laplacian} $L(x,y)$ can be computed from an image $I(x,y)$ by convolving it with the the kernel
\begin{equation}
    \mathbf{D}_{xy}^2 = \begin{pmatrix}
        0 & 1 & 0 \\
        1 & -4 & 1 \\
        0 & 1 & 0
    \end{pmatrix}
\end{equation}
as
\begin{equation}
\nabla^2 I(x,y) \equiv L(x,y) = I(x,y)*\mathbf{D}_{xy}^2,
\end{equation}
where $*$ denotes the two-dimensional convolution, and $\nabla^2$ denotes the Laplace operator. The discrete Laplacian can also be computed as the sum of two matrices
\begin{equation}
    L(x,y) = A_X(x,y) + A_Y(x,y),
\end{equation}
where
\begin{equation}
    A_X(x,y) = G_X(x,y) - G_X(x-1,y), \ \text{for $x>0$},
\end{equation}
and
\begin{equation}
    A_Y(x,y) = G_Y(x,y) - G_Y(x,y-1), \ \text{for $y>0$},
\end{equation}
where we additionally define $A_X(0,y) = G_X(0,y)$ and $A_Y(x,0) = G_Y(x,0)$ for all $x$ and $y$. Hence, we can approximate the discrete Laplacian as
\begin{equation}\label{eq:approximate_discrete_laplacian}
    \hat{L}(x,y) = \hat{A}_X(x,y) + \hat{A}_Y(x,y),
\end{equation}
where
\begin{equation}
    \hat{A}_X(x,y) = \hat{G}_X(x,y) - \hat{G}_X(x-1,y), \ \text{for $x>0$},
\end{equation}
and
\begin{equation}
    \hat{A}_Y(x,y) = \hat{G}_Y(x,y) - \hat{G}_Y(x,y-1), \ \text{for $y>0$},
\end{equation}
with  $\hat{A}_X(0,y) = \hat{G}_X(0,y)$ and $\hat{A}_Y(x,0) = \hat{G}_Y(x,0)$ for all $x$ and $y$ as boundary conditions. Notice that the resolution compressed approximation of the discrete Laplacian can be computed by replacing the quantized gradients with their resolution compressed versions.
\subsubsection{Solving the Poisson's equation with successive over-relaxation}
\begin{figure}
\centering
\includegraphics[width=0.7\linewidth]{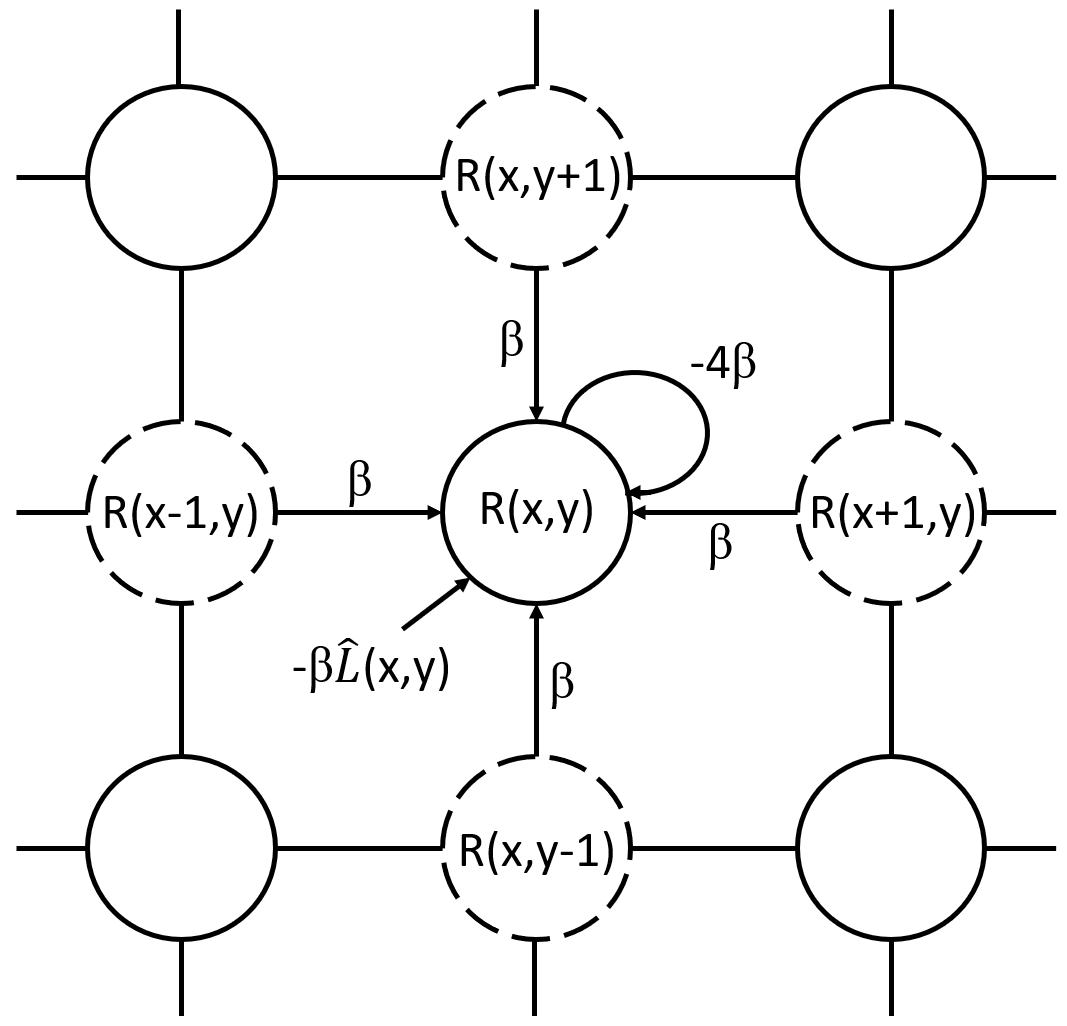}
\caption{Parallelization of the SOR algorithm. Here $R(x,y)$ denotes the reconstructed value at a given iteration step of the algorithm. Each step of iteration consists of first updating the cells visualized by solid circles, and then updating the cells visualized by dashed circles --- the lattice continues with alternating solid and dashed circles. For simplifying the notation, here the term $\beta$ is the over-relaxation parameter divided by four, $\beta = \alpha / 4$.}\label{fig:sor-visualization}
\end{figure}

Based on above we can write the following approximate version of Poisson's equation:
\begin{equation}
    \nabla^2 I(x,y) \approx \hat{L}(x,y).
\end{equation}
Poisson's equation can be solved for $I(x,y)$ by using the successive over-relaxation (SOR) algorithm originally described in~\cite{SOR}. The successive over-relaxation is an iterative method which depends on the \emph{over-relaxation} parameter $\alpha \in [1,2]$. 
A cellular architecture computing the SOR algorithm in parallel is presented in Fig.~\ref{fig:sor-visualization}. We will denote the reconstructed image by $R(x,y)$; it can be initialized for example as $R(x,y) = 0$.

\subsubsection{Reconstruction example}
\begin{figure*}
\centering
\includegraphics[width=\linewidth]{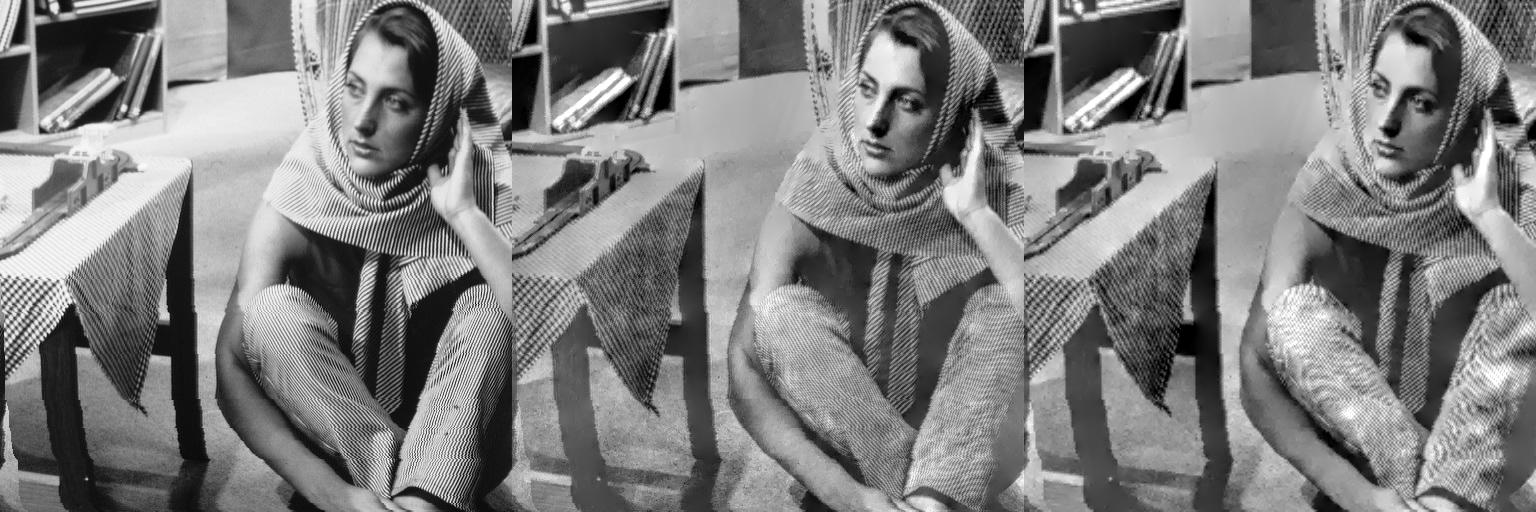}
\caption{Reconstruction from approximate Laplacian using the SOR algorithm with $\alpha=1.97$ and $k=100$ iterations. Left inset: original image. Middle inset: reconstruction using quantized gradients $\hat{G}_X(x,y)$ and $\hat{G}_Y(x,y)$. Right inset: reconstruction using resolution compressed approximate gradients $\hat{G}_X^\textrm{RC}(x,y)$ and $\hat{G}_Y^\textrm{RC}(x,y)$. The approximate Laplacians are multiplied by the constant $c=3.6$ in order to make the variance of the reconstructed image similar to the variance of the original image. For visualization, the mean values of the reconstructed images were subtracted from the reconstructed images and the mean of the original image was added to them. As can be seen, the resolution compression yields reconstruction artefacts especially in high-frequency parts of the image.}\label{fig:resolution_compression}
\end{figure*}

Reconstructed grayscale images are illustrated in Fig.~\ref{fig:resolution_compression}. The middle inset shows the result of the reconstruction from approximate gradients $\hat{G}_X(x,y)$ and $\hat{G}_Y(x,y)$ computed from ternary gradients of Fig.~\ref{fig:gradient_events}. Approximate gradients are used to compute the approximate Laplacian, which is in turn used in solving the Poisson equation using the SOR algorithm. Since the approximate gradients are smaller or equal in magnitude to the exact gradients, we have multiplied the approximate Laplacian by a empirically chosen scaling constant $c$ to make the variance of the reconstructed image close to the variance of the original image. We also zero-centered the reconstructed image and added to it the mean of the original image. As illustrated by Fig.~\ref{fig:resolution_compression}, reconstruction with three-level position-dependent quantization (using the over-relaxation parameter $\alpha = 1.97$, $k = 100$ iterations, and the scaling constant $c = 3.6$) yields a reasonably accurate estimate of the original grayscale image. This shows that ternary gradients --- and hence gradient events --- retain significant amount of information about the gradients in the original image. The right inset of Fig.~\ref{fig:resolution_compression} illustrates reconstruction with resolution compression of the ternary gradients; resolution compression results in loss of spatial details and produces some aliasing in high frequency parts of the image, but the person and the objects in the image are still easily recognizable.

A drawback of using gradient events --- and ternary gradients --- in reconstructing a grayscale image is that they contain no information of the mean intensity value of the original image. In order to visualize the results, one must add a bias value to the reconstructed image; in the reconstruction evaluations presented in this paper we use the mean value of ground truth grayscale image as this bias value. Thus we conclude, that a physical implementation of the gradient event camera would benefit from a global illumination sensor --- \emph{e.g.} a discrete photodiode --- which would then yield a correct value for the bias parameter. 

\section{Results}\label{sec:results}
\begin{table*}
\caption{\label{table:results_without_b} Reconstruction comparison against different event-based reconstruction methods presented in the literature, scored by the metrics defined in~\cite{evreal}. The reconstructed images are zero-centered, and the mean value of the ground truth image is added to them as the gradient events do not contain information on the mean intensity. The best and second best scores are presented in \textbf{bold} and \underline{underlined}. Grad. events and Grad. events (RC) refer to results obtained with gradient events and the successive over-relaxation reconstruction method without and with resolution compression, respectively. The following reconstruction parameters were used for gradient events: $(\alpha = 1.97, k = 100, c = 3.6)$.}
\centering
\begin{tabular}{lccccccccc}
\toprule
 & \multicolumn{3}{c}{Dataset: ECD} & \multicolumn{3}{c}{MVSEC} & \multicolumn{3}{c}{HQF}\\
\midrule
 & MSE$\downarrow$ & SSIM$\uparrow$ & LPIPS$\downarrow$ & MSE$\downarrow$ & SSIM$\uparrow$ & LPIPS$\downarrow$ & MSE$\downarrow$ & SSIM$\uparrow$ & LPIPS$\downarrow$ \\
E2VID~\cite{E2VID} & 0.019 & 0.657 & 0.209 & 0.076 & 0.257 & 0.598 & 0.036 & 0.530 & 0.343 \\
Firenet~\cite{Firenet} & 0.008 & 0.732 & 0.163 & 0.037 & 0.369 & 0.538 & 0.041 & 0.488 & 0.408 \\
E2VID+~\cite{HQF} & 0.008 & 0.700 & 0.156 & 0.030 & 0.388 & 0.414 & 0.024 & 0.576 & 0.235 \\
FireNet+~\cite{HQF} & 0.017 & 0.541 & 0.273 & 0.054 & 0.284 & 0.516 & 0.031 & 0.490 & 0.316 \\
SPADE-E2VID~\cite{SPADE-E2VID} & 0.008 & 0.675 & 0.207 & 0.035 & 0.400 & 0.478 & 0.050 & 0.440 & 0.483 \\
SSL-E2VID~\cite{SSL-E2VID} & 0.021 & 0.528 & 0.336 & 0.064 & 0.306 & 0.636 & 0.055 & 0.448 & 0.473 \\
ET-Net~\cite{ET-Net} & 0.010 & 0.679 & 0.184 & 0.035 & 0.371 & 0.438 & 0.027 & 0.544 & 0.268 \\
HyperE2VID~\cite{evreal_webpage} & 0.010 & 0.660 & 0.190 & 0.035 & 0.366 & 0.436 & 0.025 & 0.538 & 0.264 \\
Grad. events & \underline{\textbf{0.002}} & \textbf{0.850} & \textbf{0.068}  & \underline{0.017} & \textbf{0.676} & \textbf{0.212} & \underline{0.013} & \textbf{0.815} & \textbf{0.125}\\
Grad. events (RC) & \underline{\textbf{0.002}} & \underline{0.828} & \underline{0.091} & \textbf{0.016} & \underline{0.606} & \underline{0.255} & \textbf{0.012} & \underline{0.781} & \underline{0.159} \\
\toprule
\end{tabular}
\end{table*}

The following datasets were used in the quantitative evaluation of this work: the Event-Camera Dataset (ECD)~\cite{ECD}, the Multi Vehicle Stereo Event Camera Dataset (MVSEC)~\cite{MVSEC} and the High Qualify Frames (HQF)~\cite{HQF} dataset. We modified the updated version of the event-to-video evaluation pipeline~\cite{evreal, evreal_webpage} to accommodate the evaluation of the gradient event -based reconstruction of grayscale images.

\begin{figure}
\centering
\includegraphics[width=\linewidth]{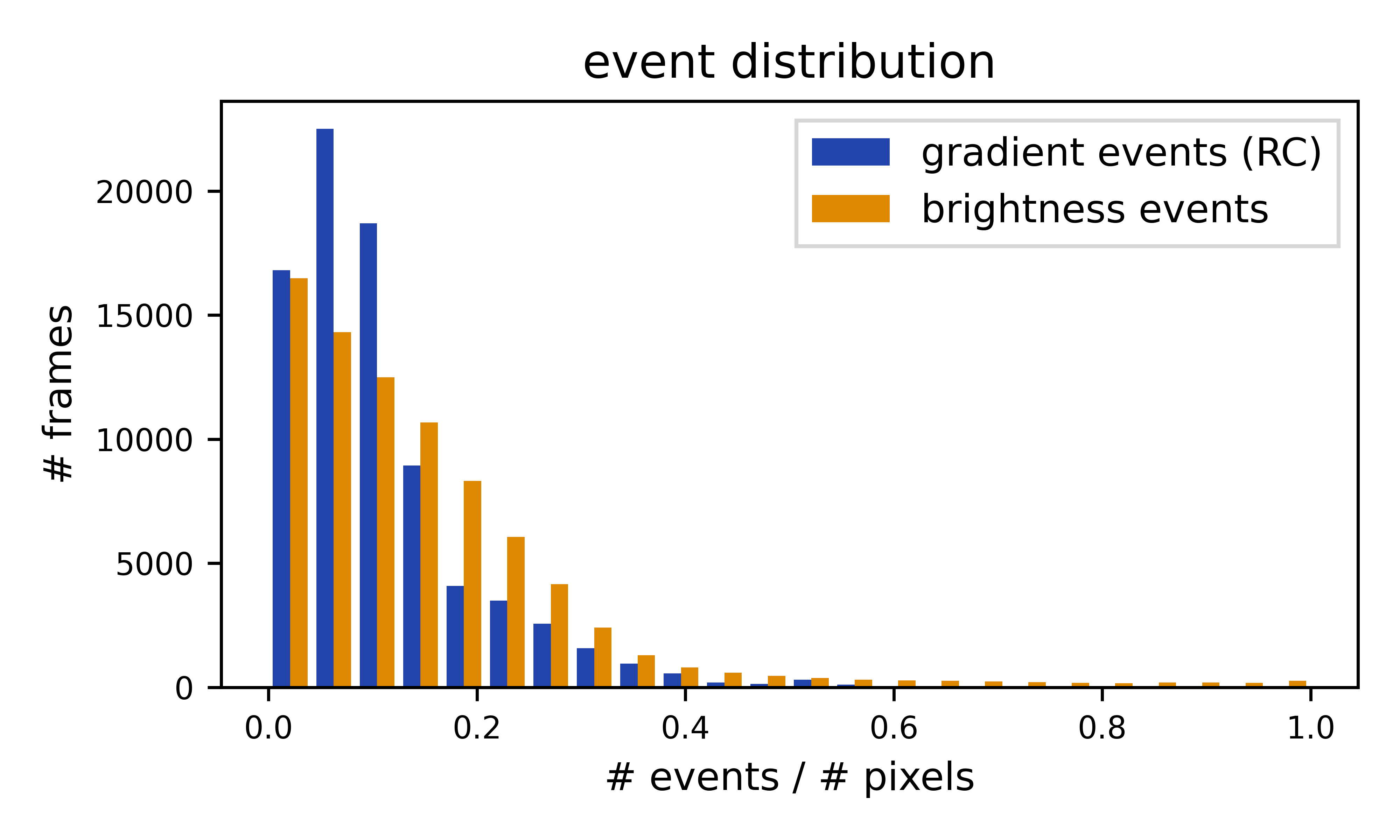}
\caption{Number of frames corresponding to event probability (the number of events normalized by the number of pixels in a frame) over the datasets ECD, MVSEC and HQF, corresponding to the ground truth grayscale frames. For the brightness events, at most one event per pixel position was counted, and the count was obtained for timestamps between the previous and the considered ground truth frame. For the gradient events, resolution compression was used in order to have the spatial resolution the same as for brightness events. The total event probability for gradient events with resolution compression was approximately 0.11, and the total event probability for brightness events was approximately 0.15.}\label{fig:event_distributions}
\end{figure}

Gradient events (with and without resolution compression) were computed from the grayscale frames of the videos using the pixel position -dependent quantization thresholds $\{t_0 = 4/255, \ t_1 = 8/255, \ t_2 = 16/255\}$. Fig.~\ref{fig:event_distributions} illustrates event count distributions (normalized by the number of pixels per frame) for brightness and gradient events with resolution compression. For these distributions we counted the number of pixel positions with at least one brightness event for each ground truth frame (for each ground truth frame we considered the brightness events whose timestamps were between the time of the previous frame and the time of the considered frame), while the number of gradient events was simply obtained by counting the number of non-zero events generated from the corresponding ground truth frame.

With the selected quantization thresholds, the overall probability $p_g$ of a gradient event without resolution compression (averaged over all frames and pixel positions) is approximately equal to twice of the overall probability $p_b$ of a brightness event ($p_g \approx 0.3$, $p_b \approx 0.15$) in the considered datasets. Thus on average, the probability of a single type of gradient event (horizontal or vertical) is approximately equal to the probability of a brightness event. When resolution compression is used (and both types of gradient events are taken into account), the overall probability of a gradient event is $p_{g,RC} \approx 0.11$, which is less than the probability of a brightness event $p_b \approx 0.15$.

Reconstruction of grayscale images from gradient events (with or without resolution compression) was performed as explained in Section~\ref{subsec:reconstruction} with the reconstruction parameters ($\alpha = 1.97$, $k = 100$, $c = 3.6$). The reconstructed images were made to have the same mean value as the ground truth frames, since the gradient events contain no information of the grayscale mean intensity value. Brightness event -based video reconstruction methods presented in publications~\cite{E2VID, Firenet, HQF, SPADE-E2VID, SSL-E2VID, ET-Net, HyperE2VID} were evaluated using the event processing pipeline~\cite{evreal}. The reconstructed frames were also made to have the same mean value as the corresponding ground truth frames in order to have a fair comparison against the gradient event -based reconstructions. Additionally, for the E2VID~\cite{E2VID} and SSL-E2VID~\cite{SSL-E2VID} methods, the so-called robust min/max normalization was used as a post-processing step. These two reconstruction methods tend to output reconstructed images with quite small ranges of intensity values and hence their evaluation scores would be significantly worse than the scores for the other methods, if this normalization were not applied.

The reconstruction evaluation results over the considered three datasets are presented in Table~\ref{table:results_without_b}: in all of the quantitative evaluations the gradient event reconstructions (with or without resolution compression) performed significantly better than any of the brightness event -based reconstruction methods. We conclude that while the gradient and brightness event probabilities are similar, the reconstruction quality using gradient events is clearly superior. A potential explanation for this is that gradient events capture more information of the visual scene than brightness events do.

\section{Discussion}\label{sec:discussion}

A gradient event camera would have several advantages when compared to a conventional event camera. As noted in~\cite{tobi_center_surround}, conventional event cameras suffer from the generation of unwanted events in the presence of fluctuating illumination resulting from the use of e.g. LEDs and fluorescent lamps. A gradient event camera operating in logarithmic domain, on the other hand, would be less sensitive to flickering lights: ternary gradients represent differences between intensity values of neighboring pixels, and thus the additive term corresponding to global illumination is cancelled.

Since ternary gradients can be used to approximate the Laplacian of an image, reconstruction can be performed using well-known Poisson equation solvers, such as the SOR algorithm. This shows an additional advantage of gradient events: the reconstruction algorithm contains only three adjustable parameters (the over-relaxation parameter $\alpha$, number of iterations $k$, and the scaling parameter $c$) in contrast to the deep networks used in reconstructing frames from brightness events, which contain tens of thousands to millions of parameters, as presented in Table~\ref{table:number_of_parameters}.

\begin{table}
\caption{\label{table:number_of_parameters}Number of adjustable parameters in by different event reconstruction methods (partially adopted from~\cite{evreal}). E2VID+ and FireNet+ have the same number of adjustable parameters as E2VID and FireNet, respectively.}
\begin{tabular}{lr}
\toprule
Reconstruction method & Number of adjustable parameters\\
\midrule
E2VID~\cite{E2VID} & $10.71\times 10^6$\\
FireNet~\cite{Firenet} & $40\times 10^3$\\
SPADE-E2VID~\cite{SPADE-E2VID} & $11.46\times 10^6$\\
SSL-E2VID~\cite{SSL-E2VID} & $57\times 10^3$ \\
ET-Net~\cite{ET-Net} & $22.18\times 10^6$\\
HyperE2VID~\cite{evreal_webpage} & $10.15\times 10^6$\\
This work (SOR) & \textbf{3}\\
\toprule
\end{tabular}
\end{table}

\begin{figure*}
\centering
\includegraphics[width=\linewidth]{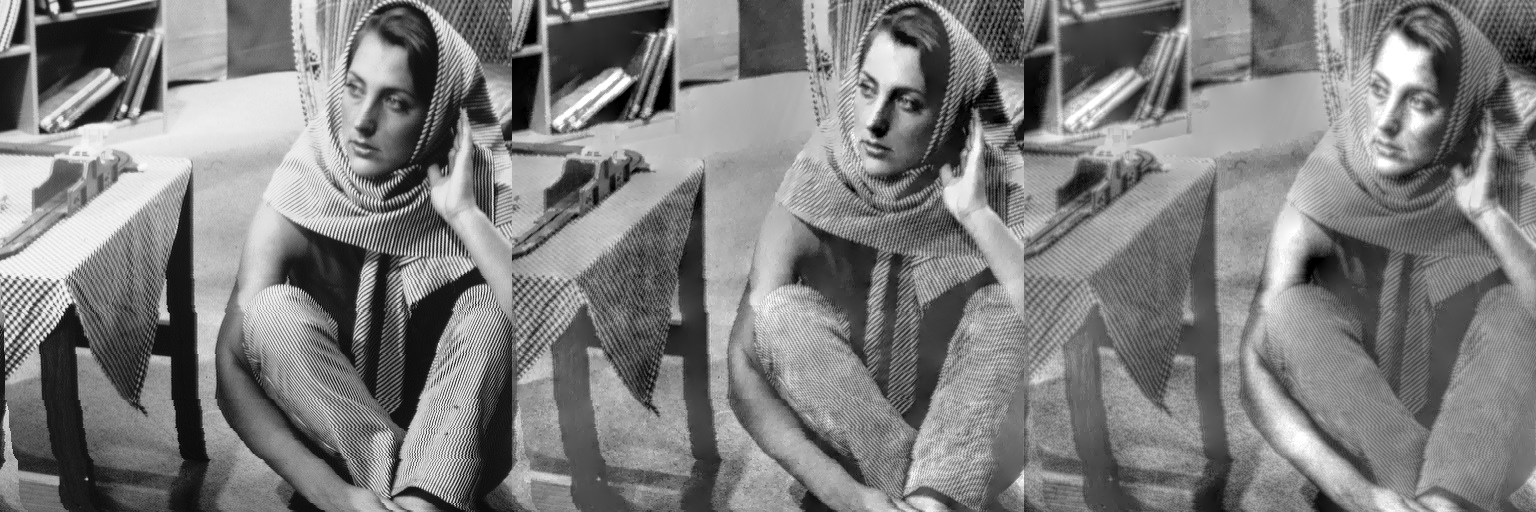}
\caption{Effect of using multiple thresholds on the reconstruction quality. Left inset: original image. Middle inset: thresholds $\{4/256, 8/256, 16/256\}$ and scaling parameter $c=3.6$. Right inset: threshold $4/256$ and scaling parameter $c=4.3$. There are more reconstruction artefacts for example around the head and the left hand when using only one threshold as compared to using three thresholds.}\label{fig:three_vs_one_thresholds}
\end{figure*}

The quality of gradient event reconstruction is significantly higher when the position-dependent threshold approach is used as opposed to using only one threshold. If, for example, one would use only one threshold $4/255$ instead of the set of thresholds $\{4/255, 8/255, 16/255\}$, the reconstructed grayscale image would be much noisier. It is beneficial to detect different magnitudes of gradients even when it comes with the price of having lower sampling frequency of any given gradient magnitude. This is illustrated in Fig.~\ref{fig:three_vs_one_thresholds}, where the middle inset is obtained with set set of three threshold values using position-dependent thresholding, and the right inset is obtained with only one threshold value. 

\begin{figure*}
\centering
\includegraphics[width=\linewidth]{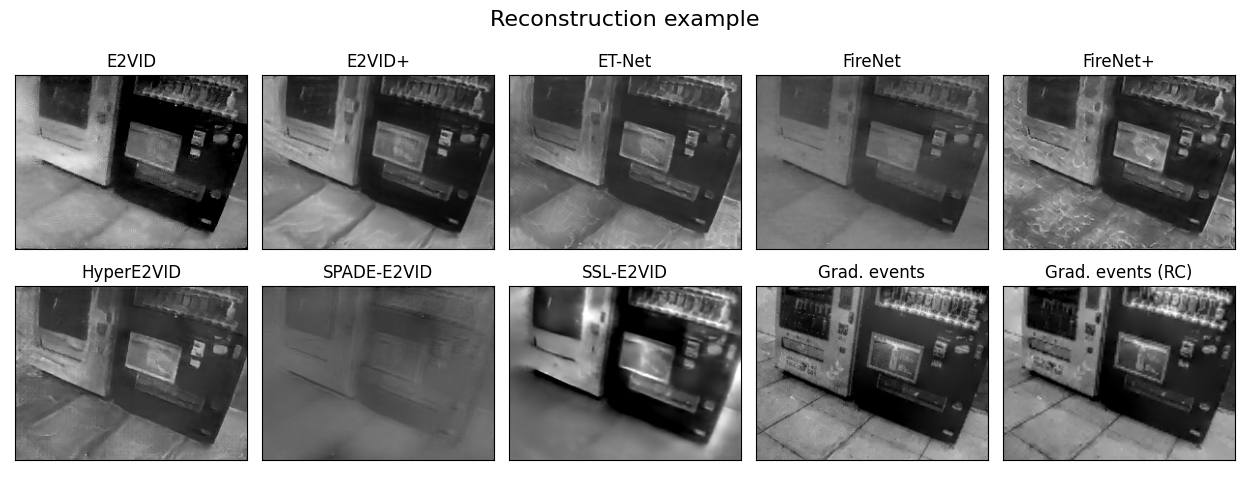}
\caption{Reconstructed images from the bike\_bay\_hdr video from the HQF~\cite{HQF} dataset.}\label{fig:geometry_example}
\end{figure*}

We note that geometrical properties of objects are typically preserved with better accuracy using gradient event -based reconstruction as compared to brightness event -based reconstruction. An example of this is presented in Fig.~\ref{fig:geometry_example}, which presents corresponding frames from reconstructed videos. It can be seen that the gradient event -based reconstruction methods retain more details of the scene when compared to brightness event -methods, and do not yield curvy artefacts such as those visible in E2VID+ and ET-Net reconstructions.

A shortcoming of the presented analysis is that the gradient events are not obtained from a physical implementation of a gradient event camera. This may raise the question on the fairness of the quantitative comparison between the reconstruction methods. However, we argue that a physical gradient event camera can work in similar fashion as the presented emulation: measured grayscale pixel intensities are turned into ternary gradients and further into gradient events, and for example the noise present in the grayscale pixel values affects the quality of the ternary gradients. We also note that the overall event probabilities presented in Fig.~\ref{fig:event_distributions} show that on average, the brightness event reconstruction methods have access to a larger number of events than the resolution compressed gradient event reconstruction method. In future we will perform quantitative comparison between brightness event and gradient event -based reconstruction methods using a physical implementation of a gradient event camera.

\section{Conclusion}
In this work we proposed a new gradient event that could be used in future generations of event-based cameras. We assessed the information content of gradient events by evaluating the quality of reconstruction of grayscale images. Our conclusion is that gradient event -based reconstruction yields higher quality grayscale images than methods based on conventional brightness events. Additional benefit of using gradient events is that the reconstruction method does not introduce any inherent lag in the reconstructed images, as the reconstruction is based only on the latest event values. Furthermore, the reconstruction can be implemented using the simple SOR algorithm, which has only three adjustable parameters. The SOR algorithm can be efficiently computed on a GPU, as it is based on updating cells which communicate only between nearest neighbors.

Gradient events offer a hardware-friendly and efficient compression method of visual information, and should also be useful in various downstream computer vision applications. We believe that the presented work serves as a fruitful starting point for an interesting new line of research in neuromorphic engineering.

\section*{Acknowledgements}
This project has received funding from the European Union's Horizon 2020 research and innovation programme under grant agreement no 101016734.

\bibliographystyle{IEEEtran}
\bibliography{references}

\end{document}